%% file: iclr2024_conference.tex
\documentclass{article} 
\usepackage{iclr_set,times}

\input{math_commands.tex}

\usepackage{hyperref}
\usepackage{url}
\usepackage{graphicx}
\usepackage{bbm}
\usepackage{booktabs} 
\usepackage{tabularx}
\usepackage{colortbl}
\usepackage{xcolor}
\definecolor{graycell}{gray}{0.9}
\usepackage{arydshln}
\usepackage{ulem}
\usepackage{tablefootnote}
\usepackage{subfigure}
\usepackage{array}

\usepackage{floatrow}
\newfloatcommand{capbtabbox}{table}[][\FBwidth]
\usepackage{wrapfig}
\usepackage{float}
\usepackage{caption}
\usepackage{multirow}

\title{Large Language Model Bias Mitigation from the Perspective of Knowledge Editing}

\iclrfinalcopy

\author{
Ruizhe Chen \\
Zhejiang University  \\ \And Yichen Li\ $^\dagger$ \\
Zhejiang University \\
\And Yang Feng \\
Angelalign Technology Inc.\And Zuozhu Liu \thanks{Corresponding author. $^\dagger$ Equal Contribution} \\
Zhejiang University \\
}

\begin{document}

\maketitle

\begin{abstract}
Existing debiasing methods inevitably make unreasonable or undesired predictions as they are designated and evaluated to achieve parity across different social groups but leave aside individual facts, resulting in modified existing knowledge.
In this paper, we first establish a new bias mitigation benchmark BiasKE leveraging existing and additional constructed datasets, which systematically assesses debiasing performance by complementary metrics on fairness, specificity, and generalization. Meanwhile, we propose a novel debiasing method, Fairness Stamp (FAST), which enables editable fairness through fine-grained calibration on individual biased knowledge. Comprehensive experiments demonstrate that FAST surpasses state-of-the-art baselines with remarkable debiasing performance while not hampering overall model capability for knowledge preservation, highlighting the prospect of fine-grained debiasing strategies for editable fairness in LLMs.
\end{abstract}

\section{Introduction}

Pre-trained Large Language Models (LLMs) have demonstrated exceptional performance on many tasks~\citep{devlin2018bert, floridi2020gpt, brown2020language}. However, the encoded social stereotypes and human-like biases inevitably cause undesired behaviors when deploying LLMs in practice~\citep{zhao2019gender, navigli2023biases, sheng2021societal}.
Existing approaches to mitigate biases in LLMs are mainly categorized into: (1) Fine-tuning~\citep{zmigrod2019counterfactual, webster2020measuring, he2022mabel, liang2020towards, lauscher2021sustainable}, which includes techniques such as re-balanced corpus pre-training, contrastive learning, projection methods, and efficient parameter tuning. (2) Prompt-tuning~\citep{guo2022auto, yang2023adept, li2023prompt, dong2023co}, which involves creating prompts to address social biases.

\begin{figure*}[htb]

\vspace{-0.1in}
	\centering  
		\includegraphics[width=1.0\linewidth]{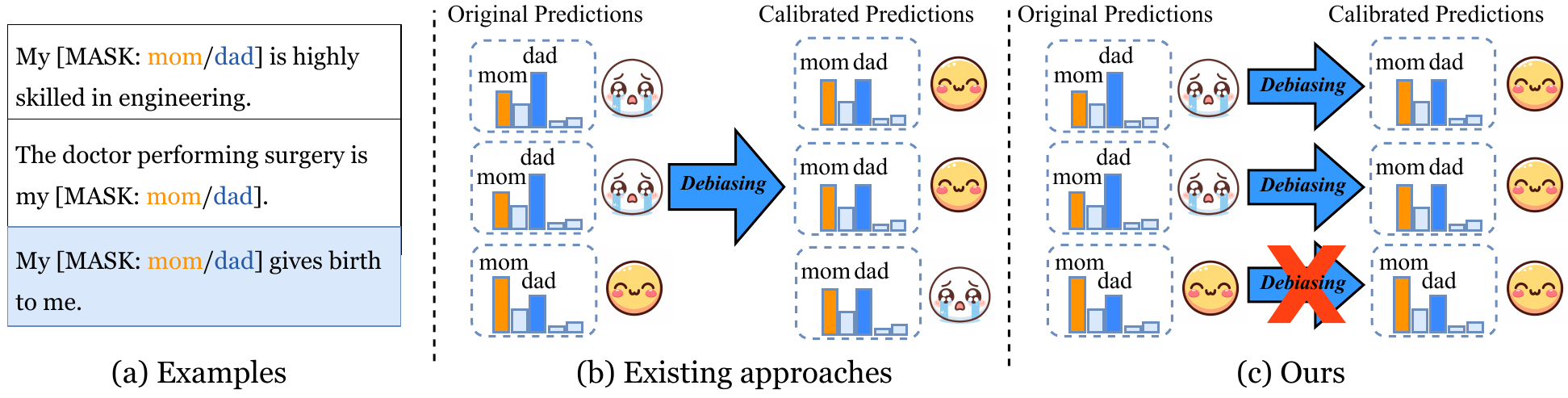}
	\caption{\small (a) Expression towards different groups (e.g., mom/dad) does not necessarily constitute a bias. (b) Existing debiasing approaches usually equalize different groups, resulting in unreasonable predictions. (c) Our proposed method performs fine-grained calibration with biased knowledge, while maintaining the others.}
    \label{fig:Illustration}
\end{figure*}

However, existing techniques treat social groups as interchangeable~\citep{gallegos2023bias} and neutralize protected attributes of different social groups in model inputs or outputs, while ignoring or concealing distinct mechanisms of different social groups~\citep{hanna2020towards}, as shown in Figure~\ref{fig:Illustration}.
Furthermore, existing debiasing evaluation metrics mainly focus on the degree of bias, but fail to measure whether the model retains its origin knowledge~\citep{gallegos2023bias} of discerning reasonable disparities among different social groups.

To address these issues, we first establish a more comprehensive debiasing benchmark \textbf{BiasKE} by extending existing datasets with additional constructed data and evaluation metrics on fairness, specificity, and generalization.
Moreover, we propose a novel method Fairness-Stamp (\textbf{FAST}) for editable bias mitigation. Instead of mitigating group biases indiscriminately, FAST operates fine-grained calibrations on individual biases, i.e., specific stereotyped statements toward a social group.
Specifically, we first design a causal-tracing-based method to locate the decisive layer in LLMs responsible for biased predictions. Then we propose to add a lightweight modular network, which enables fine-grained and efficient debiasing of one or multiple individual biased knowledge, with objectives of bias mitigation and knowledge maintenance. 

We evaluate FAST with comprehensive experiments on StereoSet~\citep{stereoset2020} and Crows-Pairs~\citep{nangia2020crows}, which are further extended as BiasKE for systematic evaluation. Results show that FAST achieves remarkable debiasing performance without compromising model capability.
We extend FAST to larger models such as GPT-Neo and Llama to demonstrate the scalability in real-world applications.
Additional experiments showcase the effectiveness on downstream tasks, continual bias mitigation, and lightweight optimization, with results and analysis in Appendix~\ref{more Analysis}.

\section{BiasKE Benchmark Construction}

\begin{figure*}[htb]
	\centering  
		\includegraphics[width=1\linewidth]{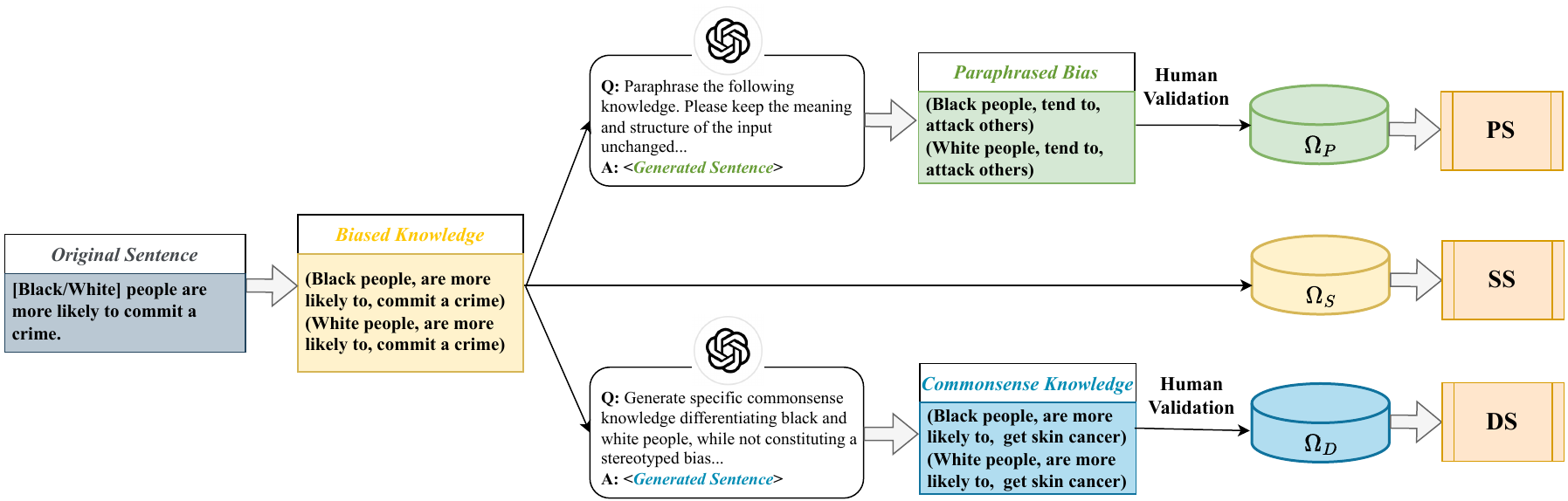}
	\caption{An illustration of the construction of BiasKE.}
        \label{illustration of construction of BiasKE}
\end{figure*}

In this section, we describe the procedures for establishing BiasKE, with an illustration in Figure~\ref{illustration of construction of BiasKE}.
To better express a bias, we formalize the stereotype bias (e.g., Man is good at man) as a triplet $k = (s, r, o)$, where $s$ is the subject (i.e., Man), $o$ is the object (i.e., math), and $r$ is the relation between them (i.e., is good at), as inspired by \citet{petroni2019language}.
We collect social biases related to three domains (gender, race, and religion) from six existing datasets, as detailed in Appendix~\ref{Biased Knowledge Benchmark dataset}. 

\textbf{Step1.} Based on these social biases, we extract biased knowledge pairs ${(k_1, k_2)}$. As shown in Figure~\ref{illustration of construction of BiasKE}, the sentence ``\textit{black people are more likely to commit a crime}'' can be extracted as $k_1$ \textit{(Black people, are more likely to, commit a crime.)}. $k_2$ is the counterfactual of $k_1$, which can have an opposite $s_2$ (i.e., white people) or $o_2$ (i.e., compliance). Representative examples of different datasets can be referred to in Table~\ref{constructing biased knowledge pair from different datasets.}. The set of biased knowledge pairs is denoted by $\Omega_{S}$.


\textbf{Step2.} Then we create \(\Omega_{P}\), the set of paraphrased biased knowledge pair $(k_1^{'}, k_2^{'})$, with the same semantic expression as ${k_1, k_2}$, as exemplified in Figure~\ref{illustration of construction of BiasKE}.
\(\Omega_{P}\) constitutes similar social biases as in $\Omega_{S}$, which is utilized to measure the generalization ability of debiased models and prevent the edited model from overfitting to a particular input.

\textbf{Step3.} Finally, \(\Omega_{D}\) is independently created by collecting commonsense knowledge related to the subjects (e.g., \textit{man/woman}, \textit{Christians/Jewish}) in $\Omega_{S}$. We also confirm that pre-existing knowledge in \(\Omega_{D}\) is irrelevant to the knowledge within \(\Omega_{S}\), thus measuring the ability to retain unrelated knowledge.
Both \(\Omega_{P}\) and \(\Omega_{D}\) are initially generated by prompting GPT-4 API and manually validated.

\textbf{Evaluating Metrics.} Furthermore, for fair and systematic evaluation, we design three evaluating metrics, Stereotype Score (SS), Paraphrase Stereotype Score and Differentiation Score (DS), to evaluate fairness, generalization and specificity ability of debiasing methods, respectively. Specifically, in addition to using SS to measure the degree of bias, PS evaluates the generalization ability on semantically similar biased knowledge, and DS
evaluates the ability to preserve existing knowledge about
individuals. Detailed descriptions of these evaluating metrics are presented in Appendix~\ref{Biased Knowledge Benchmark}.

\section{Method}

\begin{figure*}[htb]
	\centering  
	\includegraphics[width=0.9\linewidth]{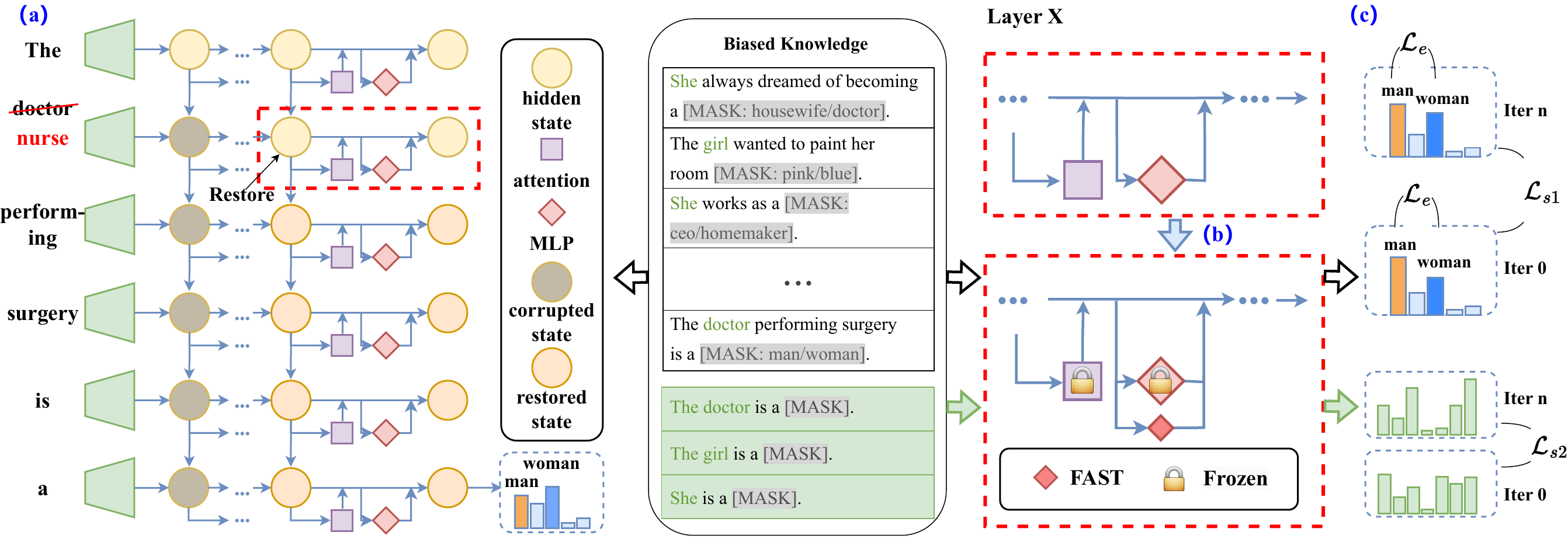}
	\caption{\small An illustration of our FAST framework. (a) We first localize the critical layer towards biased predictions. (b) A fairness stamp is inserted within the critical layer. (c) Our FAST can finely calibrate debiasing demands with the objective of bias mitigation and knowledge maintenance.}
    \label{fig:pipeline}
\end{figure*}


We propose a fine-grained bias mitigation method Fairness-Stamp (\textbf{FAST}). 
FAST operates through a two-step process, as depicted in Figure~\ref{fig:pipeline}.
In the first step, we propose to investigate if there are specific hidden states (i.e., layers) that play a more crucial role than others when recalling biased knowledge, as inspired by the knowledge localization works~\citep{meng2022locating,finlayson2021causal}.
Our biased knowledge localization is performed in three steps, biased run, counterfactual input and restoration run, with a complete description in Figure~\ref{fig:Locate Biased Knowledge} in the Appendix~\ref{Locate Biased Knowledge app}:

In the second step, we propose to select the layer that contributes most significantly to the bias and envelope it with a Fairness Stamp. The fairness stamp is a 2-layer Feed-Forward Network (FFN) layer, which adjusts the output of the enveloped layer with the same input. Assuming the input hidden states to be $\mathbf{h}$, the FFN layer in original LLMs can be formulated as follows: $\text{FFN}(\mathbf{h}) = \text{Act}(\mathbf{hK}^\top) \mathbf{V},$
where $\mathbf{K}$ and $\mathbf{V}$ denote the parameters (i.e., keys and values matrices) of the first and second linear layers in the FFN, respectively. Our fairness stamp inserts an extra intervention on the original output with a few external parameters. The new output of the modified FFN layer is: 
\begin{equation}
\text{FFN}'(\mathbf{h}) = \text{FFN}(\mathbf{h}) + \text{Act}(\mathbf{hK'}^\top) \mathbf{V'},
\end{equation}
where $\mathbf{K'}$, $\mathbf{V'}$ $\in$ $R^{d_c\times d}$ are the new parameter matrices in our fairness stamp. The stamp is optimized for each individual biased knowledge in the set $\Omega$ with the objectives of fairness (i.e., bias mitigation) and specificity (i.e., knowledge maintenance).

\textbf{Fairness.} The main objective is to mitigate the biased prediction. 
With prompts of a biased knowledge pair, we narrow the gap between predictions on the biased object and unbiased object:
\begin{equation}
\label{Efficacy}
    \mathcal{L}_e = \frac{1}{\lvert \Omega \rvert}\sum_{(k_1, k_2)\in \Omega} \lvert  \mathcal{P}_{\mathcal{G}}[k_1]- \mathcal{P}_{\mathcal{G}}[k_2]\rvert,
\end{equation}
where $k_i = (s_i,r_i,o_i)$ and $\mathcal{P}_{\mathcal{G}}[k_i] = \mathcal{P}_{\mathcal{G}}[o_i|p_i]$ denotes the probability of predicting $o_i$ given the prompt $p_i = (s_i, r_i)$.  

\textbf{Specificity.} 
We propose to preserve existing knowledge in two parts. First, we maintain the predictions for the input prompts on other objects. Furthermore, we minimize the change of predictions on simple prompts $p'$ (e.g., ``\textit{\{subject\} is a [MASK]}''), which helps preserve the perception of the model on the subjects (e.g., \textit{man, woman}). The two losses are formulated as follows:
\begin{align}
    &\mathcal{L}_{s1} = \frac{1}{\lvert \Omega \rvert}\sum_{p_i\in \Omega}\mathcal{D}_{KL} (\mathcal{P}_{\mathcal{G}}[\star|p_i], \mathcal{P}_{\mathcal{G}^{*}}[\star|p_i]),\ 
    &\mathcal{L}_{s2} = \frac{1}{\lvert \Omega \rvert}\sum_{s_i\in \Omega}\mathcal{D}_{KL} (\mathcal{P}_{\mathcal{G}}[\star|p'(s_i)], \mathcal{P}_{\mathcal{G}^{*}}[\star|p'(s_i)]),
\end{align}
where $\mathcal{P}_{\mathcal{G}}[\star|p']$ is the predicted probability vector. $\mathcal{G}$ and $\mathcal{G}^{*}$ represent the origin and debiased model.
$\mathcal{D}_{KL}$ represents the Kullback-Leibler Divergence.
To prevent the model from overfitting to particular inputs, we also utilize prefix texts \( x_j \) to enhance generalization ability across various contexts. These prefix texts are randomly generated by the model, for instance, “\textit{My father told me that}”, and are concatenated to the front of the prompts.  

The overall objective is formulated as: $\mathcal{L} = \mathcal{L}_e + \alpha \mathcal{L}_{s1} + \beta \mathcal{L}_{s2},$
where $\alpha$ and $\beta$ are hyper-parameters.

\section{Experiment}
\label{Experiment}

\textbf{Experimental Details.}
Experiments are mainly conducted on \textbf{BERT}~\citep{devlin2018bert} and \textbf{GPT2}~\citep{radford2019gpt2} compared with 8 state-of-the-art baselines. We also conduct additional experiments on larger models, i.e., GPT2-XL, GPT-Neo, and Llama-2 to further validate the scalability of FAST.
We evaluate \textbf{SS}, \textbf{PS}, \textbf{DS}, \textbf{LMS}, and \textbf{ICAT} for comprehensive comparison, with detailed description in the Appendix~\ref{Biased Knowledge Benchmark}.
We report results on \textbf{StereoSet}~\citep{stereoset2020} and \textbf{Crows-Pairs}~\citep{nangia2020crows} datasets to keep consistent with baselines.  
Details of datasets, baselines, model and implementation are reported in Appendix~\ref{more Experiment details}.
We only report the experimental results in terms of gender, please refer to the Appendix~\ref{more Debiasing Results} for race and religion.

\begin{wraptable}{r}{0.6\textwidth}
\vspace{-3pt}
\centering
\vspace{-5pt}
\scalebox{0.77}{
\renewcommand{\arraystretch}{1.2}
\begin{tabular}{l|cccccc}
\toprule
\textbf{Method}    & $\textbf{SS}_{\textbf{S-Set}}$ $\diamond$ &$\textbf{SS}_{\textbf{Crows}}$ $\diamond$ & \textbf{PS}$\diamond$ & \textbf{DS}$\uparrow$    & \textbf{LMS}$\uparrow$   & \textbf{ICAT}$\uparrow$ \\
\midrule
\cellcolor{graycell}BERT & \cellcolor{graycell}60.28     & \cellcolor{graycell}57.25     & \cellcolor{graycell}59.17            & \cellcolor{graycell}100.0 & \cellcolor{graycell}84.17 & \cellcolor{graycell}68.11 \\
CDA & 59.61 & 56.11 & 57.56 & 75.00 & 83.08 & 70.11 \\
Dropout & 60.68 & 55.34 & 58.65 & 87.50 & 83.04 & 66.95 \\
INLP & 56.66 & 51.15 & 54.15 & 66.67 & 80.63 & 71.40 \\
SelfDebias & 59.34 & 52.29 & 57.45 & 68.75 & 84.09 & 69.92 \\
SentDebias & 59.37 & 52.29 & 56.78 & 70.83 & 84.20 & 69.56 \\
MABEL & 56.25 & 50.76 & 54.74 & 66.67 & 84.54 & 73.98 \\
AutoDebias & 59.65 & 48.43 & 57.64 & 58.33 & 86.28 & 69.64 \\
FMD & 57.77 & - & 55.43 & 70.83 & 85.45 & 72.17 \\
\midrule
\textbf{Ours} & \textbf{51.16} & \textbf{49.69} & \textbf{50.80} & \textbf{95.83} & \textbf{86.30} & \textbf{84.29} \\
\bottomrule
\end{tabular}
}
\caption{\small Debiasing Results on BERT. The best result is indicated in \textbf{bold}. $\diamond$: the closer to 50, the better. ``-'': results are not reported.}
\label{Debiasing Results on BERT}
\vspace{-15pt}
\end{wraptable}

\textbf{Debiasing Results on BERT.}
The results are reported in Table~\ref{Debiasing Results on BERT}.
It is observed that all baseline methods fail to yield satisfactory results in knowledge maintenance (i.e., DS). This proves our claim that group-invariant methods compromise the ability to distinguish between different social groups while mitigating biases. However, our FAST can largely maintain a high DS.
Furthermore, our FAST is the first to achieve near-perfect bias mitigation (i.e., SS), while SS of all baselines are still higher than 56 as for StereoSet. 
This demonstrates the effectiveness of our FAST towards eliminating social biases in LLMs. 

\textbf{Debiasing Results on GPT2.}
As for GPT2, our method can consistently surpass all the baselines in terms of SS and DS, indicating its superiority in both bias mitigation and knowledge maintenance, as shown in Table~\ref{debiasing gender GPT2}. FAST also enhances the ICAT score from 68.74 to 80.38, exceeding the second-best result by 6.86. 
More debiasing results and qualitative study can be referred to Appendix~\ref{More exp}.

\textbf{Scalibility to Larger Models.}
The results on large models are reported in Table~\ref{Debiasing Results on larger models}.
After debiasing, FAST induces a significant reduction in SS, and a great improvment in ICAT. Meanwhile, FAST can also largely maintain the differentiation score for larger language models. These demonstrate the consistent effectiveness of FAST on LLMs and scalability in real-world applications. 

More analysis and discussion on language modeling capability, knowledge locating, computational complexity and hyper-parameters are provided in the Appendix~\ref{more Analysis}.

\begin{minipage}{1\textwidth}
    \begin{minipage}[t]{0.5\textwidth}
        \centering
        
        \captionof{table}{\small Debiasing Results on GPT2.}
        \vspace{-5pt}
        \scalebox{0.63}{
        \renewcommand{\arraystretch}{1.3}
        \begin{tabular}{l|cccccc}
        \toprule
        \textbf{Method}    & $\textbf{SS}_{\textbf{S-Set}}$ $\diamond$ &$\textbf{SS}_{\textbf{Crows}}$ $\diamond$ & \textbf{PS}$\diamond$ & \textbf{DS}$\uparrow$    & \textbf{LMS}$\uparrow$   & \textbf{ICAT}$\uparrow$ \\
        \midrule
        \cellcolor{graycell}GPT2 & \cellcolor{graycell}62.65 & \cellcolor{graycell}56.87 & \cellcolor{graycell}60.26 & \cellcolor{graycell}100.0 & \cellcolor{graycell}91.01 & \cellcolor{graycell}68.74 \\
        CDA & 64.02 & 56.87 & 61.12 & 67.86 & 90.36 & 65.02 \\
        Dropout & 63.35 & 57.63 & 64.29 & 71.00 & \textbf{90.40} & 64.44 \\
        INLP & 59.83 & 53.44 & 57.78 & 60.71  & 73.76 & 61.38 \\
        SelfDebias & 60.84 & 56.11 & 58.97 & 64.29 & 89.07 & 70.72 \\
        SentDebias & 56.05 & 56.11 & 57.67 & 71.43 & 87.43 & 73.52 \\
        \midrule
        \textbf{Ours} & \textbf{54.91} & \textbf{51.62} & \textbf{53.83} & \textbf{82.14} & 89.42 & \textbf{80.38} \\
        \bottomrule
        \end{tabular}
        }
        \label{debiasing gender GPT2}    
    \end{minipage}
    \begin{minipage}[t]{0.5\textwidth}
        \centering
        \captionof{table}{\small Debiasing Results on larger models.}
        \vspace{-5pt}
        \scalebox{0.63}{
        \renewcommand{\arraystretch}{1.42}
        \begin{tabular}{lcccccc}
        \toprule
        \textbf{Method}    & $\textbf{SS}_{\textbf{S-Set}}$ $\diamond$ &$\textbf{SS}_{\textbf{Crows}}$ $\diamond$ & \textbf{PS}$\diamond$ & \textbf{DS}$\uparrow$    & \textbf{LMS}$\uparrow$   & \textbf{ICAT}$\uparrow$ \\
        \midrule
        \textbf{GPT2-XL}       & 68.70           & 65.41       & 64.35            & 100.0 & 92.79            & 58.09 \\ 
        \textbf{Ours}          & 60.50           & 50.94       & 56.89            & 85.71  & 89.14            & 70.42 \\ \midrule
        \textbf{GPT-Neo}  & 70.40           & 63.52       & 68.23            & 100.0 & 93.47            & 55.33 \\ 
        \textbf{Ours}          & 60.97           & 50.96       & 60.34            & 90.48  & 84.49            & 65.95 \\ \midrule
        \textbf{Llama-2}    & 66.28           & 65.41       & 66.16            & 100.0 & 88.83            & 59.92 \\ 
        \textbf{Ours}          & 55.70           & 51.57       & 54.79            & 78.57  & 86.89            & 76.98 \\ 
        \bottomrule
        \end{tabular}
        }
        \label{Debiasing Results on larger models}
    \end{minipage}
\end{minipage}

\section{Conclusion}
In this paper, we pioneer the fine-grained bias mitigation paradigm, which specifically focuses on human-relevant individual social biases/facts rather than broad group differences. We develop a novel evaluation benchmark BiasKE and propose the first Editable Fairness framework, FAST, capable of mitigating single social biases and scalable to mitigating thousands of biases concurrently. Extensive experiments across various models and datasets demonstrate the efficacy of our approach, showcasing its generalizability, specificity, and scalability. Our findings offer significant implications for future debiasing research. The limitation and future works can be referred to Appendix~\ref{Limitation and Future Works}.



\newpage
\normalem
\bibliography{iclr2024_conference}
\bibliographystyle{iclr2024_conference}

\newpage
\appendix
\section{BiasKE Benchmark Construction}

\subsection{Metrics}
\label{Biased Knowledge Benchmark}

\textbf{Stereotype Score (SS)} is the most straightforward measure for the \textbf{bias} within the debiased model~\citep{nadeem2020stereoset, nangia2020crows}.
It computes the percentage of knowledge for which a model assigns the biased object as opposed to the unbiased object. The evaluation of \textbf{SS} is conducted according to the following criteria:
\begin{align}
\label{eqn:Bias}
 \textbf{SS}({\mathcal{G}^*},\Omega_{S}) = \mathbb{E}_{(k_1,k_2) \in \Omega_{S}}\mathbbm{1}\{\mathcal{P}_{\mathcal{G}^*}[k_1] > \mathcal{P}_{\mathcal{G}^*}[k_2]\},
\end{align}
where $\mathcal{G}^*$ is the debiased model.

\textbf{Paraphrase Stereotype Score (PS)} indicates the ability to \textbf{generalize} the learned knowledge to fairly predict on similar or related knowledge in $\Omega_{P}$. It also computes the percentage of knowledge that a model gives a biased prediction as opposed to an unbiased prediction:
\begin{align}
\label{eqn:Gen}
\textbf{PS}(\mathcal{G}^*, \Omega_{P}) =  \mathbb{E}_{(k_1^{'}, k_2^{'}) \in \Omega_{P}}\mathbbm{1}\{\mathcal{P}_{\mathcal{G}^*}[k_1^{'}] > \mathcal{P}_{\mathcal{G}^*}[k_2^{'}]\}.
\end{align}

\textbf{Differentiation Score (DS)} indicates the \textbf{specificity} of the debiasing process, which quantifies the percentage of pre-existing commonsense knowledge in $\Omega_{D}$ retained after debiasing.
The evaluation of \(\textbf{DS}\) is conducted according to the following criteria:
\begin{align}
\label{eqn:Spe}
\textbf{DS}(\mathcal{G}, {\mathcal{G}^*},\Omega_{D}) =  \mathbb{E}_{k \in \Omega_{D}}\mathbbm{1}\{\mathcal{P}_{\mathcal{G}}[k] = \mathcal{P}_{\mathcal{G}^*}[k]\}.
\end{align}

\textbf{Language Modeling Score (LMS)}, employed in StereoSet~\citep{nadeem2020stereoset}, has been adopted to further evaluate the debiasing specificity. Based on the knowledge pairs in \(\Omega_{S}\), we select an irrelevant \(o_{ir}\) to form \(k_{ir} = (s, r, o_{ir})\). LMS represents the percentage that a model that prefers a relevant association (either the stereotypical association or the anti-stereotypical association) as opposed to an irrelevant association. The evaluation of \(\textbf{LMS}\) is conducted according to the following criteria:

\begin{align}
\label{eqn:LMS}
\textbf{LMS}({\mathcal{G}},\Omega_{S}) =  \mathbb{E}_{(k_1, k_2) \in \Omega_{S}}\mathbbm{1}\{\mathcal{P}_{\mathcal{G}}[k_1] > \mathcal{P}_{\mathcal{G}}[k_{ir}]\}  +\mathbbm{1}\{\mathcal{P}_{\mathcal{G}}[k_2] > \mathcal{P}_{\mathcal{G}}[k_{ir}]\}.
\end{align}

\textbf{Ideal Context Association Test Score (ICAT)} is proposed by \citep{stereoset2020} combine both LMS and SS by $\text{ICAT} = \text{LMS}* \text{min}(\text{SS},100-\text{SS})/50$. It represents the language modeling ability of a model while behaving in an unbiased manner.

\subsection{Dataset.}
\label{Biased Knowledge Benchmark dataset}
We collect biased knowledge related to three domains (gender, race, and religion) from six existing datasets (StereoSet~\citep{nadeem2020stereoset}, Crows-Pairs~\citep{nangia2020crows}, WEAT~\citep{caliskan2017semantics}, WinoBias~\citep{zhao2018gender}, Winogender~\citep{rudinger2018gender} and BEC-Pro~\citep{bartl2020unmasking}). These datasets have been benchmarked to detect biases within Language Models (LLMs). The statistics of our constructed knowledge base can be referred to Table~\ref{The statistics of collected biased knowledges}, with a detailed description referred to in the following.

\textbf{StereoSet}~\citep{nadeem2020stereoset} employs a methodology to evaluate a language model's propensity for stereotypical associations. The procedure is essentially a fill-in-the-blank challenge, where the model is given a sentence with a missing word and must select from a stereotypical word, an anti-stereotypical word, or an irrelevant word. 

\textbf{CrowS-Pairs}~\citep{nangia2020crows} constitutes a dataset featuring intrasentential minimal pairs. Each pair comprises one sentence depicting a socially disadvantaged group in a manner that either conforms to or contradicts a stereotype, and another sentence that is slightly altered to reference a contrasting, advantaged group. The language model's task involves assessing the probability of masked tokens that are exclusive to each sentence within these pairs.

\textbf{WEAT}~\citep{caliskan2017semantics} is comprised of word sets that pertain to either attributes or targets. It evaluates the associations between concepts of social groups (for instance, masculine and feminine terms) and neutral attributes (such as terms related to family and occupation).

\textbf{Winogender}~\citep{rudinger2018gender} and \textbf{Winobias}~\citep{zhao2019gender} are designed to assess gender-based stereotypical associations with various occupations. In some instances, these evaluations involve associating gender-specific pronouns with occupations that are stereotypically linked to that gender. In other cases, the task is to associate pronouns with occupations that are typically considered non-stereotypical for that gender.

\textbf{BEC-Pro} (The Bias Evaluation Corpus with Professions)~\citep{bartl2020unmasking} is a tool for assessing gender biases in the context of occupations. It comprises 5,400 sentences, each generated from a template that includes a term denoting a person and one of 60 professional terms. During the evaluation process, both the person-related and professional words in these sentences are masked for analysis.

\subsection{Dataset Construction}
\textbf{Paraphrased dataset.} For each knowledge pair within \(\Omega_{S}\), we paraphrase the prompts combining \((s, r)\) with the same semantic expression. 
We hired 2 undergraduate students, all with good English proficiency. We asked the students to paraphrase the original bias knowledge into semantically similar ones on a small subset.
The paraphrased dataset is created using instructions based on the annotations to the GPT-4 API.

\textbf{Differentiation dataset.}
We construct \(\Omega_{D}\) by collecting commonsense facts related to the sensitive attributes, such as ``\textit{Jesus' resurrection is commemorated by [Christians] when they celebrate Easter.}'' We first create alternative facts by prompting GPT-4 API, then a manual validation process is employed to ensure that \(\Omega_{D}\) contains reasonable facts. We also manually confirm that \(\Omega_{D}\) does not contain the same knowledge as \(\Omega_{S}\). 

\begin{table}[h]
\centering
\caption{The statistics of collected biased knowledge in our BiasKE. ``-'' means not included.}
\vspace{2mm}
\scalebox{0.9}{
\begin{tabular}{lccc}
\toprule
\textbf{Source/domain} & \textbf{gender} & \textbf{race} & \textbf{religion} \\ 
\midrule
StereoSet & 771 & 2976 & 247 \\
Crows-Pairs & 262 & 516 & 105 \\
WEAT\tablefootnote{WEAT contains attribute word sets and target word sets that embed biased correlations. In this Table, we count the total number of attribute words.} & 128 & 188 & 18 \\
WinoBias\tablefootnote{WinoBias, Winogender, and BEC-Pro model biased correlations between gender and occupation. We categorize these data under the gender domain.} & 1584 & - & - \\
Winogender & 60 & - & - \\
BEC-Pro & 5400 & - & - \\
\bottomrule
\end{tabular}
}
\label{The statistics of collected biased knowledges}
\end{table}

\begin{table*}[h]
\centering
\caption{Examples of constructing biased knowledge pair from different datasets.}
\vspace{2mm}
\small
\begin{tabular}{lll}
\toprule
\textbf{Source} & \textbf{Case Sample} & \textbf{Biased Knowledge Pair} \\ 
\midrule
StereoSet & \scriptsize The people of Afghanistan are [MASK: violent/caring/fish]. & \scriptsize (Afghanistan, are, violent), (Afghanistan, are, caring) \\
\midrule
Crows-Pairs & \scriptsize My [dad/mom] spent all day cooking for Thanksgiving & \scriptsize(dad, spent all day, cooking), (mom, spent all day, cooking) \\
\midrule
WEAT & \scriptsize Christian (from Attribute Word Set 1), peace (Target Word Set 1) & \scriptsize (Christian, loves, peace), (Christian, loves, sickness) \\
\midrule
WinoBias & \scriptsize[The developer] argued with the designer because [she] did... & \scriptsize(developer, argued..., she), (developer, argued..., he) \\
\midrule
Winogender & \scriptsize The technician told the customer that she could pay with cash.& \scriptsize (technician, told..., she), (technician, told..., he) \\
\midrule
BEC-Pro &\scriptsize He is a bus mechanic. &\scriptsize (He, is a, bus mechanic), (She, is a, bus mechanic) \\
\bottomrule
\end{tabular}
\label{constructing biased knowledge pair from different datasets.}
\end{table*}

\section{Method}

\subsection{Locate Biased Knowledge}
\label{Locate Biased Knowledge app}
\begin{figure*}[htb]
	\centering  
		\includegraphics[width=1.0\linewidth]{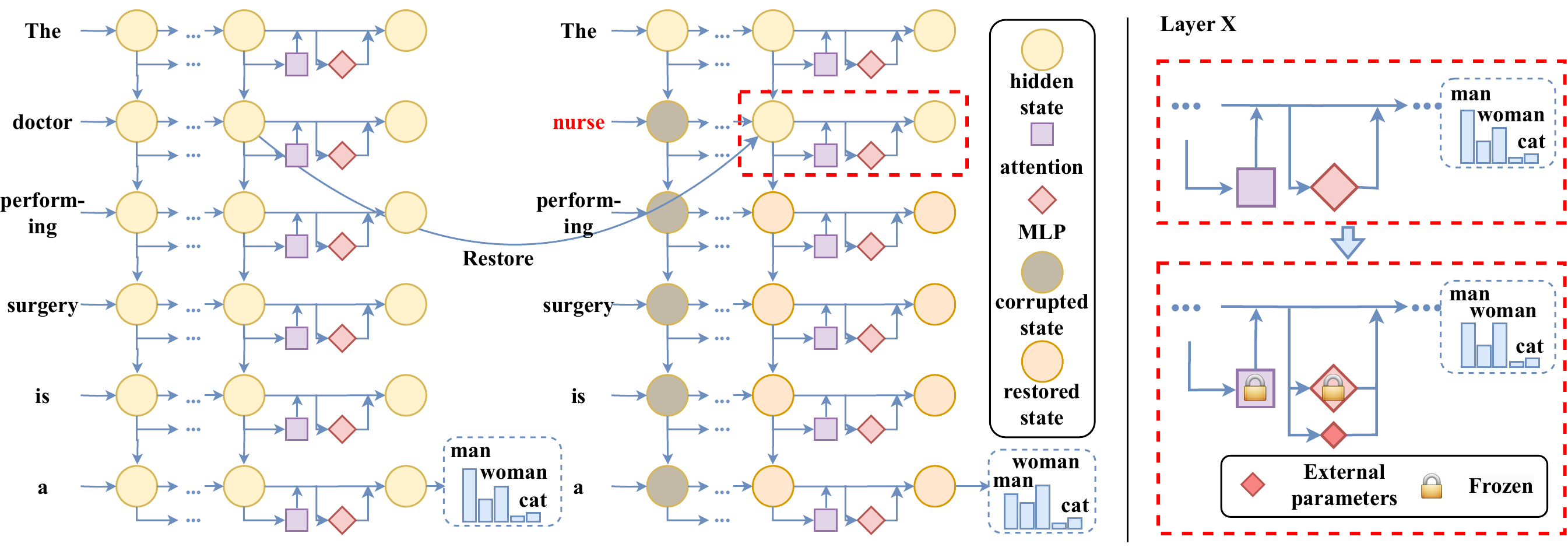}
	\caption{Illustration of our debiasing framework.}
    \label{fig:Locate Biased Knowledge}
\end{figure*}

In this section, we provide a complete illustration of our Step~1 in Figure~\ref{fig:Locate Biased Knowledge}. 

Denote $(s_1, r, o)$ as a biased knowledge such as (\textit{The doctor, performing surgery is a, man}). $(s_2, r, o)$ is the counterfactual knowledge (i.e., $s_2$ is \textit{The nurse}).
Our biased knowledge localization is performed in three steps, with a complete illustration in Figure~\ref{fig:Locate Biased Knowledge} in the Appendix:

\textbf{Biased run}: We pass the prompt $(s_1, r)$ into the model and collect all hidden states $\{h^{(l)}_i$ $|$ $i \in [1, T]$, $l \in [1, L]\}$ where $T$ is number of tokens and $L$ is number of layers.

\textbf{Counterfactual input}: We replace the subject with $s_2$ and pass the new prompt $(s_2, r)$ to the model to corrupt the biased prediction. Hidden states corresponding to the subject token(s) $\hat{i}$ will be updated with $h_{\hat{i}}^{(0)} (s_1 \rightarrow s_2)$. 

\textbf{Restoration run}: Towards certain layer $\hat{l}$ in the model, we hook the biased states $h_{\hat{i}}^{(\hat{l})}$ at subject token(s) $\hat{i}$ and perform the counterfactual run. Then we calculate the recovery degree of biased prediction, which indicates the causal effect of $\hat{l}$ to biased prediction. The layer with highest causal effect will be selected as the decisive layer.

\textbf{Causal effect.} Denote $\mathcal{P}[o]$, $\mathcal{P}^{*}[o]$ as the probability of biased prediction and counterfactual prediction. Let $\mathcal{P}^{*}(h_{\hat{i}}^{(\hat{l})})[o]$ denotes the probability of counterfactual prediction with restoration of the biased states $h_{\hat{i}}^{(\hat{l})}$. The indirect causal effect (IE) of a certain layer can be calculated by $\text{IE} = \mathcal{P}^{*}(h_{\hat{i}}^{(\hat{l})})[o] - \mathcal{P}^{*}[o]$.

\section{Experiment}
\label{More exp}

\subsection{Experiment details}
\label{more Experiment details}

\textbf{Baselines.}
We consider the following debiasing techniques as baselines. The techniques can be grouped into two categories. \textit{(1) Fine-tuning}: \textbf{Counterfactual Data Augmentation (CDA)}\footnote{We use the reproduction of CDA, Dropout, SentenceDebias, INLP and Self-Debias provided by \url{https://github.com/McGill-NLP/bias-bench}}~\citep{zmigrod2019counterfactual} involves re-balancing a corpus by swapping bias attribute words (e.g., he/she) in a dataset. The re-balanced corpus is then often used for further training to debias a model. \textbf{Dropout}~\citep{webster2020measuring} proposes to increase the dropout parameters and perform an additional phase of pre-training to debias. \textbf{SentenceDebias}~\citep{liang2020towards} proposes to obtain debiased representation by subtracting biased projection on the estimated bias subspace from the original sentence representation. \textbf{Iterative Nullspace Projection (INLP)}~\citep{ravfogel2020null} is also a projection-based debiasing technique to remove protected property from the representations. \textbf{MABEL}\footnote{We use the debiased models provided in \url{https://github.com/princeton-nlp/MABEL/}}~\citep{he2022mabel} mitigates Gender Bias using Entailment Labels.
\textit{(2) Prompt-tuning}: \textbf{Auto-debias}\footnote{We use the debiased models provided in \url{https://github.com/Irenehere/Auto-Debias}}~\citep{guo2022auto} proposes to directly probe the biases encoded in pre-trained models through prompts, then mitigate biases via distribution alignment loss. \textit{(3) Post-hoc}: \textbf{Self-Debias}~\citep{schick2021self} proposes to leverage a model’s internal knowledge to discourage it from generating biased text. 
\textbf{FMD}~\citep{chen2023fast} proposes a machine unlearning-based strategy to efficiently remove the bias in a trained model.
We also include \textbf{Fine-tuning (FT)} the original model on the same data and with the same objectives as our proposed \textbf{FAST}.

\textbf{Model.}
We mainly experiment on the representative masked language model \textbf{BERT} (\textit{bert-base-uncased})~\citep{devlin2018bert} and generative language model \textbf{GPT2} (\textit{GPT2-small})~\citep{radford2019gpt2} as our backbones. Extended experiments are also conducted on \textbf{GPT2-XL}, \textbf{GPT-Neo} (\textit{GPT-Neo-2.7b})~\citep{gpt-neo} and \textbf{Llama-2} (\textit{Llama-2-7b})~\citep{touvron2023llama}. 
We utilize pre-trained models in the Huggingface Transformers library~\citep{wolf2020transformers}.

\textbf{Implementation details.}
We utilize two-layer fully connected neural networks with the ReLU activation function as the fairness stamp. The hidden dimension is set to 1024. The batch size is set to 4. We use Adam optimizer with a learning rate of 0.1. We train each batch for 20 iterations. $\alpha$ is set to be 40 and $\beta$ is 0.1.

\subsection{Knowledge Locating Results}
\label{more Knowledge Locating Results}

\begin{figure}[htb]      

        \resizebox{0.95\textwidth}{!}{
        \subfigure{
        \includegraphics[scale=0.1]{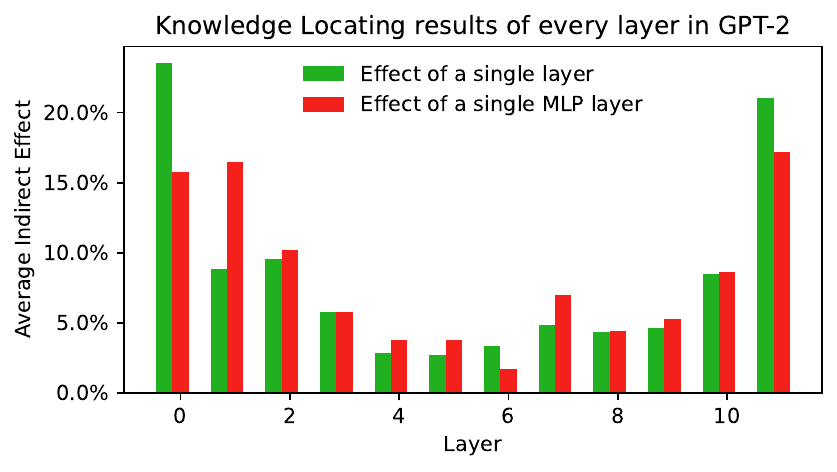}
        }
        \subfigure{
        \includegraphics[scale=0.1]{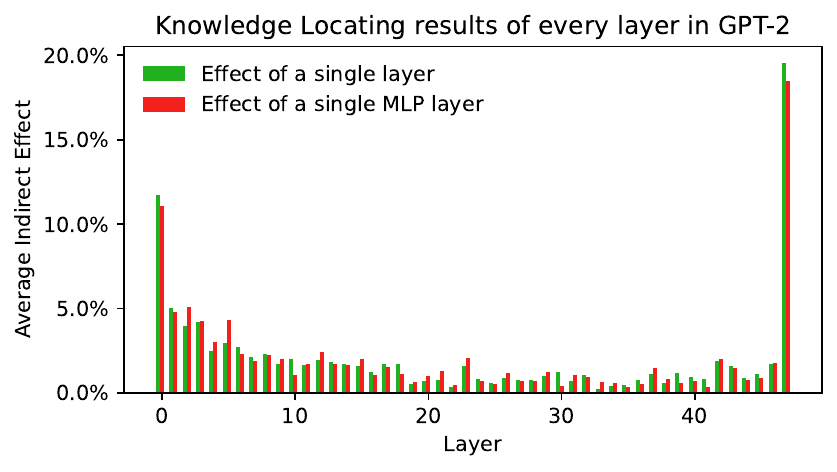}
        }
        }
        \caption{Knowledge Locating results of GPT2 (left) and GPT2-XL (right).}
        \label{Knowledge Locating results of GPT2 (left) and GPT2-XL (right).}

    \end{figure}

\begin{figure}[htb]      

        \resizebox{0.95\textwidth}{!}{
        \subfigure{
        \includegraphics[scale=0.1]{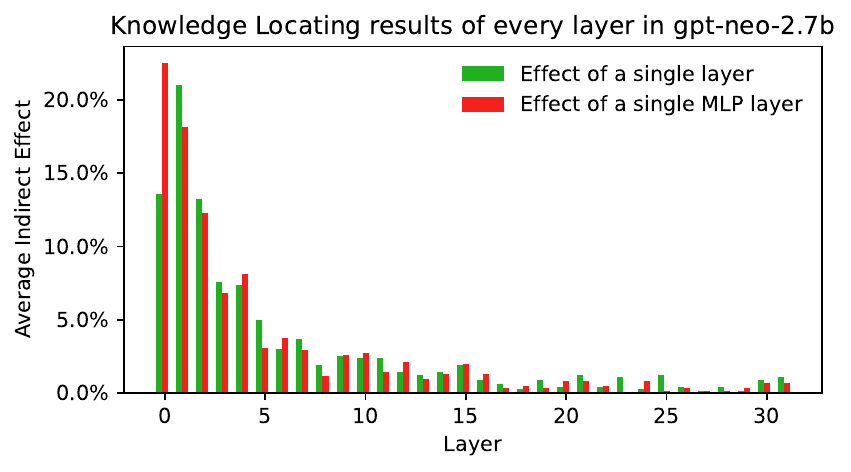}
        }
        \subfigure{
        \includegraphics[scale=0.1]{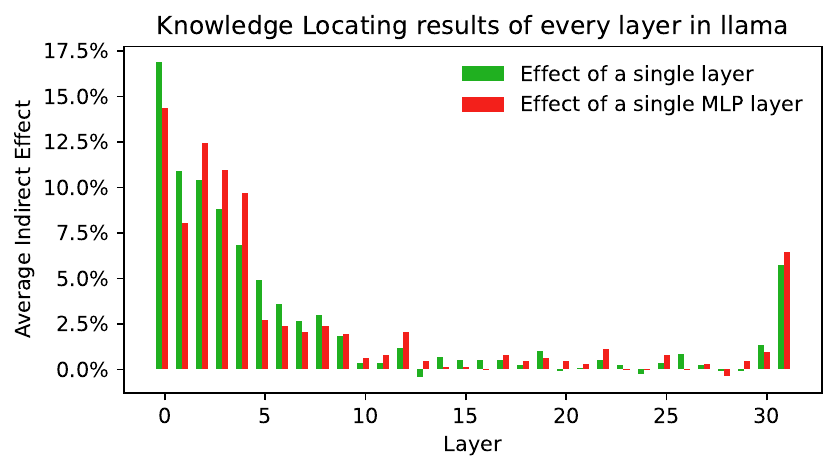}
        }
        }
        \caption{Knowledge Locating results of GPT-Neo (left) and Llama (right).}
        \label{Knowledge Locating results of GPT-Neo (left) and Llama (right).}

    \end{figure}

We present the results of knowledge locating on other backbones, as illustrated in Figure~\ref{Knowledge Locating results of GPT2 (left) and GPT2-XL (right).} and Figure~\ref{Knowledge Locating results of GPT-Neo (left) and Llama (right).}. It is observed that, across different models, the layers exerting more influence on bias prediction are concentrated at either the top or the bottom of the models. Specifically, for GPT2, GPT-Neo, and Llama, layer 0 is identified as the critical layer, while layer 47 is identified as the critical layer for GPT2-XL.

\subsection{Debiasing Results on BERT and GPT2}
\label{more Debiasing Results}

\textbf{Debiasing Results on BERT}
in terms of race and religion are supplemented in Table~\ref{Debiasing Results on BERT in terms of race and religion.}.
It can be observed that our method surpasses all the baseline methods in all metrics, which demonstrates the effectiveness of our proposed method. 

\textbf{Debiasing Results on GPT2} in terms of race and religion are presented in Table~\ref{Debiasing Results on GPT2 in term of race and religion.}, which also demonstrates the consistent performance of our method in different debiasing tasks.

\begin{table}[htbp]
\centering
\caption{Debiasing Results on BERT in terms of race and religion. $\diamond$: the closer to 50, the better. The best result is indicated in \textbf{bold}.}
\vspace{2mm}
\scalebox{0.71}{
\begin{tabular}{l|cccccc|cccccc}
\toprule
\textbf{Attribute}  & \multicolumn{6}{c|}{\textbf{Race}}                                      & \multicolumn{6}{c}{\textbf{Religion}}                                        \\
\midrule
\textbf{Method}    & $\textbf{SS}_{\textbf{S-Set}}$ $\diamond$ &$\textbf{SS}_{\textbf{Crows}}$ $\diamond$ & \textbf{PS}$\diamond$ & \textbf{DS}$\uparrow$    & \textbf{LMS}$\uparrow$   & \textbf{ICAT}$\uparrow$  & $\textbf{SS}_{\textbf{S-Set}}$ $\diamond$ &$\textbf{SS}_{\textbf{Crows}}$ $\diamond$ & \textbf{PS}$\diamond$ & \textbf{DS}$\uparrow$    & \textbf{LMS}$\uparrow$   & \textbf{ICAT}$\uparrow$  \\
\midrule
\cellcolor{graycell}BERT & \cellcolor{graycell}57.03 & \cellcolor{graycell}62.33 & \cellcolor{graycell}56.60 & \cellcolor{graycell}100.0 & \cellcolor{graycell}84.17 &\cellcolor{graycell}72.20 & \cellcolor{graycell}59.70 & \cellcolor{graycell}62.86 & \cellcolor{graycell}59.70 & \cellcolor{graycell}100.0 & \cellcolor{graycell}84.17 & \cellcolor{graycell}67.87 \\
CDA & 56.73 & 56.70 & 54.36 & 79.17 & 83.41 & 69.99 & 58.37 & 60.00 & 57.95 & 93.75 & 83.24 & 67.82 \\
Dropout & 56.94 & 59.03 & 55.46 & 93.75 & 83.04 & 70.84 & 58.95 & 55.24 & 59.22 & 95.83 & 83.04 & 67.90 \\
INLP & 57.36 & 67.96 & 56.89 & \textbf{100.0} & 83.12 & 70.80 & 60.31 & 60.95 & 59.59 & 97.92 & 83.37 & 65.82 \\
SelfDebias & 54.30 & 56.70 & 54.31 & 66.67 & 84.24 & 76.60 & 57.26 & 56.19 & 56.45 & 95.83 & 84.23 & 69.63 \\
SentDebias & 57.78 & 62.72 & 58.01 & 75.00 & 83.95 & 70.75 & 58.73 & 63.81 & 59.38 & 97.92 & \textbf{84.26} & 69.74 \\
MABEL & 57.18 & 56.01 & 57.11 & 75.00 & \textbf{84.32} & 72.20 & 56.15 & 52.12 & 53.54 & \textbf{100.0} & 81.95 & 71.87 \\
\midrule
\textbf{Ours} & \textbf{51.93} & \textbf{52.54} & \textbf{51.27} & 89.58 & 83.44 & \textbf{80.21} & \textbf{53.29} & \textbf{51.52} & \textbf{52.98} & \textbf{100.0} & 82.59 & \textbf{77.16} \\

\bottomrule
\end{tabular}
}
\label{Debiasing Results on BERT in terms of race and religion.}
\end{table}

\begin{table}[htbp]
\centering
\caption{Debiasing Results on GPT2 in terms of race and religion. $\diamond$: the closer to 50, the better. The best result is indicated in \textbf{bold}.}
\vspace{2mm}
\scalebox{0.71}{
\begin{tabular}{l|cccccc|cccccc}
\toprule
\textbf{Attribute}  & \multicolumn{6}{c|}{\textbf{Race}}                                      & \multicolumn{6}{c}{\textbf{Religion}}                                        \\
\midrule
\textbf{Method}    & $\textbf{SS}_{\textbf{S-Set}}$ $\diamond$ &$\textbf{SS}_{\textbf{Crows}}$ $\diamond$ & \textbf{PS}$\diamond$ & \textbf{DS}$\uparrow$    & \textbf{LMS}$\uparrow$   & \textbf{ICAT}$\uparrow$  & $\textbf{SS}_{\textbf{S-Set}}$ $\diamond$ &$\textbf{SS}_{\textbf{Crows}}$ $\diamond$ & \textbf{PS}$\diamond$ & \textbf{DS}$\uparrow$    & \textbf{LMS}$\uparrow$   & \textbf{ICAT}$\uparrow$  \\
\midrule

\cellcolor{graycell}GPT2 & \cellcolor{graycell}58.9 & \cellcolor{graycell}59.69 & \cellcolor{graycell}59.29 & \cellcolor{graycell}100.0 & \cellcolor{graycell}91.01 & \cellcolor{graycell}74.76 & \cellcolor{graycell}63.26 & \cellcolor{graycell}62.86 & \cellcolor{graycell}66.52 & \cellcolor{graycell}100.0 & \cellcolor{graycell}91.01 & \cellcolor{graycell}67.02 \\
CDA & 57.31 & 60.66 & 54.98 & 71.43 & 90.36 & 77.15 & 63.55 & \textbf{51.43} & 61.97 & \textbf{75.00} & 90.36 & 65.87 \\
Dropout & 57.5 & 60.47 & 55.21 & 75.00 & 90.40 & 76.84 & 64.17 & 52.38 & 62.84 & \textbf{75.00} & 90.4 & 64.78 \\
INLP & 55.52 & 59.69 & 59.75 & 75.00 & 89.20 & 79.47 & 63.16 & 61.90 & 62.68 & 71.43 & 89.89 & 66.33 \\
SelfDebias & 57.33 & 53.29 & 57.11 & 67.86 & 89.53 & 76.34 & 60.45 & 58.10 & 62.77 & 67.86 & 89.36 & 71.03 \\
SentDebias & 56.47 & 55.43 & 56.84 & 60.71 & \textbf{91.38} & 79.29 & 59.62 & 35.24 & 63.30 & 67.86 & \textbf{90.53} & 72.70 \\
\midrule
\textbf{Ours} & \textbf{52.35}	&\textbf{51.25}	&\textbf{52.87}	&\textbf{87.75} &	90.37 	&\textbf{86.12}  & \textbf{50.80} & 52.53 & \textbf{53.88} & \textbf{75.00} & 85.29 & \textbf{83.93} \\

\bottomrule
\end{tabular}
}
\label{Debiasing Results on GPT2 in term of race and religion.}
\end{table}

\subsection{Debiasing Results on BEC-Pro and Winogender}

We also report the debiasing performance on the test sets BEC-Pro and Winogender in Table.~\ref{Debiasing Results on BEC-Pro and Winogender.}. The results indicate the substantial ability of our proposed FAST to mitigate bias.

\begin{table}[htb]
\centering
\caption{Debiasing Results on BEC-Pro and Winogender. $\diamond$: the closer to 50, the better. The best result is indicated in \textbf{bold}.}
\vspace{2mm}
\scalebox{0.9}{
\begin{tabular}{lccccc}
\toprule
\textbf{Method}    & $\textbf{SS}_{\textbf{BEC}}$ $\diamond$ &$\textbf{PS}_{\textbf{BEC}}$ $\diamond$ & \textbf{DS}$\uparrow$ & $\textbf{SS}_{\textbf{Winogender}}$ $\diamond$ &  $\textbf{PS}_{\textbf{Winogender}}$$\diamond$ \\
\midrule
\cellcolor{graycell}BERT & \cellcolor{graycell}35.22 & \cellcolor{graycell}36.33 & \cellcolor{graycell}100.0 & \cellcolor{graycell}85.71 & \cellcolor{graycell}66.67\\

FAST                  & 50.44                    & 49.28                     & 93.75        & 52.38                    & 52.12                     \\

\bottomrule
\end{tabular}
}
\label{Debiasing Results on BEC-Pro and Winogender.}
\end{table}

\section{Analysis}
\label{more Analysis}

\subsection{Language Modeling Capability Analysis}

In this section, we evaluate our debiased models against the General Language Understanding Evaluation (GLUE) benchmark~\citep{wang2018glue} to evaluate whether language models retain their general linguistic understanding ability after bias mitigation.
As the GLUE benchmark results indicate (Table~\ref{Experimental results of GLUE tasks on BERT.}), FAST achieves better downstream performance than 5 out of 6 baselines on average, which indicates that FAST can mitigate the bias while also maintaining language modeling capability.

\begin{table*}[htb]
\centering
\caption{Experimental results of GLUE tasks on BERT. We report Matthew’s correlation for CoLA, the Spearman correlation for STS-B, and the F1 score for MRPC and QQP. For all other tasks, we report the accuracy. Reported results are means over three training runs. ``-'' means not reported. The best result is indicated in \textbf{bold} and the second best in \underline{underline}.}
\vspace{2mm}
\scalebox{0.87}{
\renewcommand{\arraystretch}{1.5}
\begin{tabular}{l|ccccccccc|c}
\toprule
\textbf{Method} & \textbf{CoLA} & \textbf{MNLI} & \textbf{MRPC} & \textbf{QNLI} & \textbf{QQP} & \textbf{RTE} & \textbf{SST} & \textbf{STS-B} & \textbf{WNLI} & \textbf{Average} \\
\midrule
\cellcolor{graycell}BERT & \cellcolor{graycell}56.78 & \cellcolor{graycell}84.76 & \cellcolor{graycell}89.54 & \cellcolor{graycell}91.51 & \cellcolor{graycell}88.06 & \cellcolor{graycell}64.62 & \cellcolor{graycell}93.35 & \cellcolor{graycell}88.24 & \cellcolor{graycell}56.34 & \cellcolor{graycell}79.24 \\
CDA & 2.07 & 84.84 & 81.22 & 84.84 & 87.85 & 47.29 & 92.32 & 40.83 & 43.66 & 62.77 \\
Dropout & 2.07 & 84.78 & 81.22 & 91.49 & 88.02 & 47.29 & 92.09 & 40.87 & 43.66 & 63.50 \\
SentDebias & 55.72 & \textbf{84.94} & 88.81 & 91.54 & 87.88 & 63.9 & \textbf{93.12} & 88.23 & \textbf{56.34} & \textbf{78.94} \\
AutoDebias & 57.01 & 84.91 & 88.54 & \textbf{91.65} & 87.92 & 64.62 & 92.89 & 88.43 & 40.85 & 77.42 \\
INLP & 56.50 & 84.78 & \textbf{89.23} & 91.38 & 87.94 & 65.34 & 92.66 & 88.73 & 54.93 & 77.05 \\
MABEL & \textbf{57.80} & 84.50 & 85.00 & 91.60 & 88.10 & 64.30 & 92.20 & \textbf{89.20} & - & - \\
\midrule
\textbf{Ours} & 55.99 & 84.75 & 87.60 & 91.47 & \textbf{88.12} & \textbf{67.15} & 92.20 & \underline{89.05} & 46.13 & \underline{78.01} \\
\bottomrule
\end{tabular}
}
\label{Experimental results of GLUE tasks on BERT.}
\end{table*}

\subsection{Knowledge Locating Results}

\begin{figure*}[htb]
  \centering
  \subfigure[]{\includegraphics[height=.25\columnwidth]{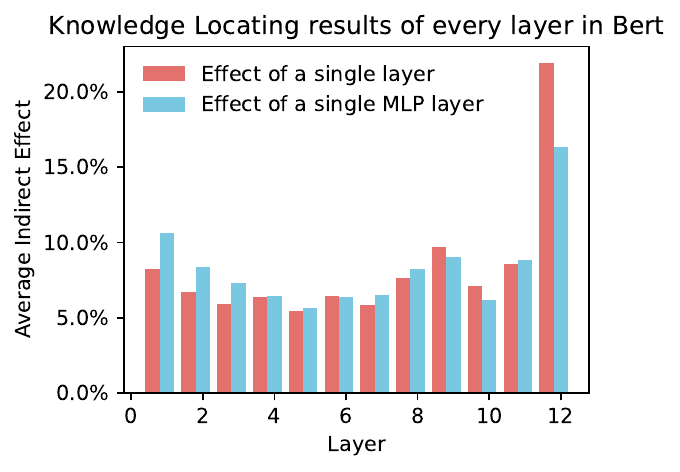}\label{Knowledge Locating results of BERT.}}
  \subfigure[]{\includegraphics[height=.25\columnwidth]{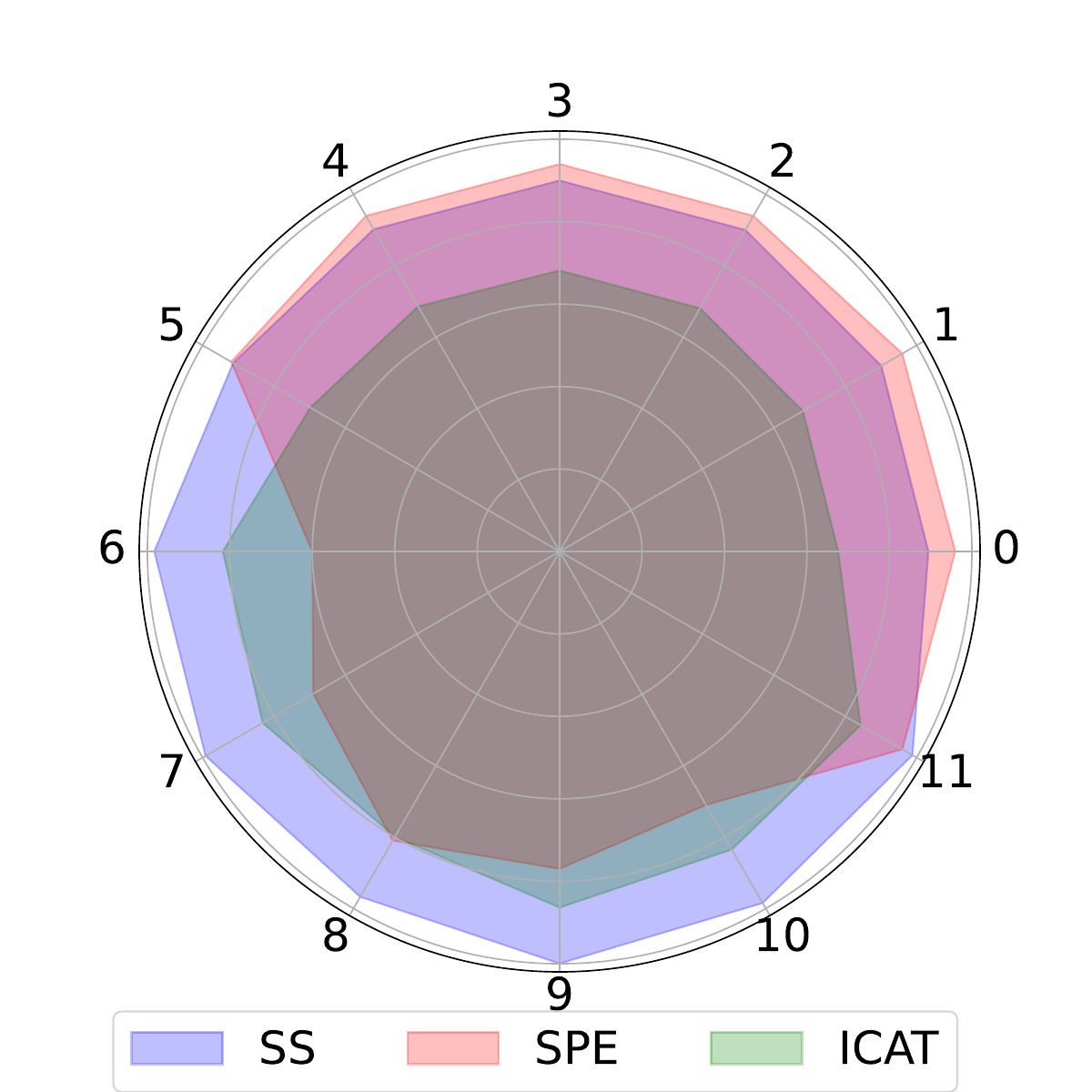}\label{Effectiveness of Knowledge Locating}}
  \subfigure[]{\includegraphics[height=.25\columnwidth]{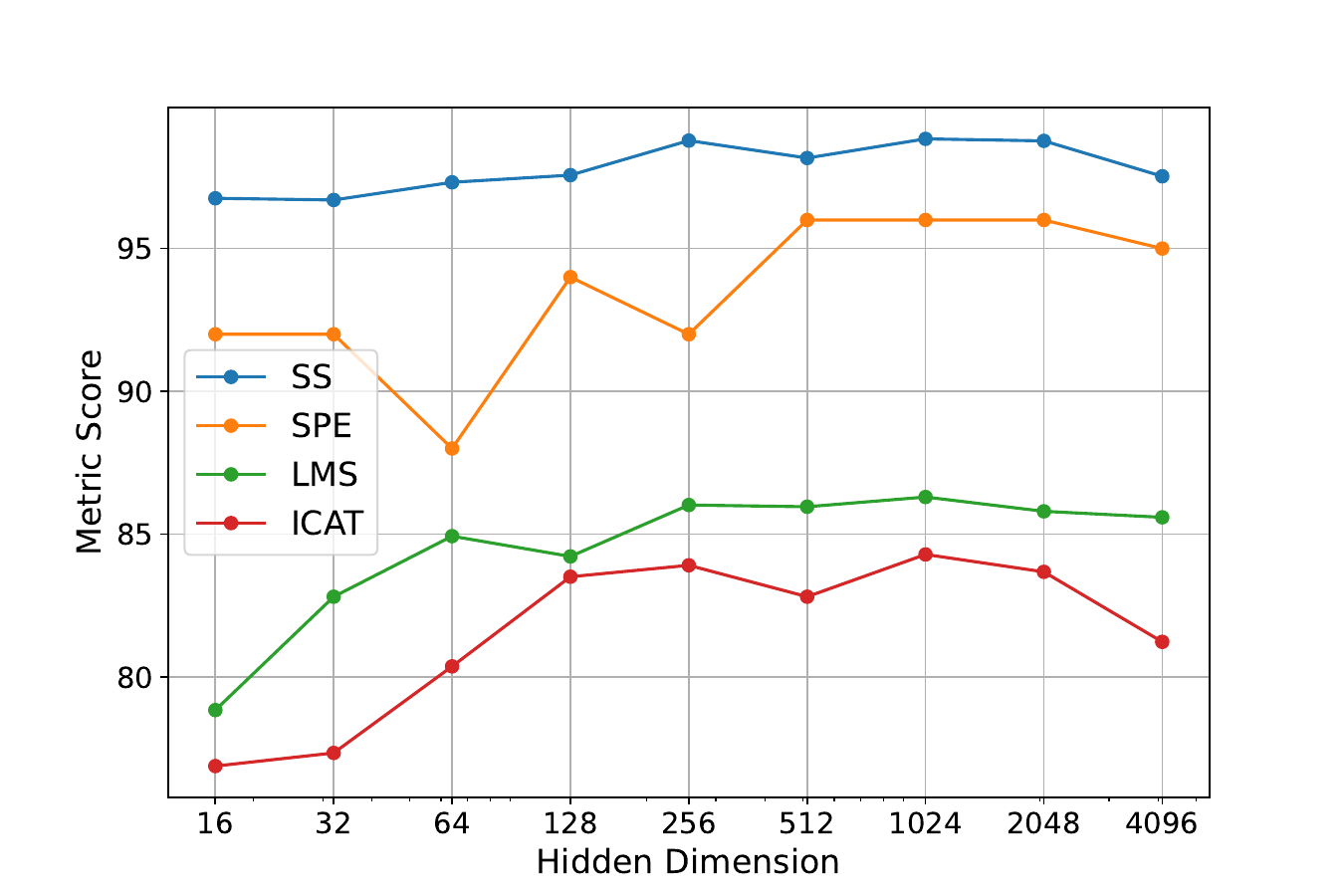}\label{Ablation on the Number of External Parameters}}
  \vspace{-0.4cm}
  \caption{(a) The average indirect effect of every layer in BERT. (b) Effectiveness verification of knowledge locating. (c) Ablation on the Number of External Parameters. Experiments are conducted on BERT in terms of gender. SS is transformed by $\text{SS}=100-|\text{SS}-50|$ so that it is also higher is better.}
\end{figure*}

In order to locate a decisive layer that contributes most to biased prediction, we separately restore each (MLP) layer in the model, and compute the average indirect effect (AIE) of different layers over the biased knowledge set. The results of BERT, as shown in Figure~\ref{Knowledge Locating results of BERT.}, reveal that the final layer of the model demonstrates an AIE significantly higher than the other layers, thus being the decisive layer of bias prediction.
In terms of GPT2, GPT2-XL, GPT-Neo, and Llama-2, as depicted in Figure~\ref{Knowledge Locating results of GPT2 (left) and GPT2-XL (right).} and Figure~\ref{Knowledge Locating results of GPT-Neo (left) and Llama (right).}, it is noticeable that the first layer contributes more significantly. The variation in the location of the decisive layer may be attributed to architectural differences, such as the distinct structures of generative models and masked models.
Detailed descriptions are reported in Appendix~\ref{more Knowledge Locating Results}.

\subsection{Effectiveness of Knowledge Locating}

To validate the effectiveness of knowledge locating (i.e., step 1 in our method), we perform calibration (i.e., step 2) on every layer of BERT, with results shown in Figure~\ref{Effectiveness of Knowledge Locating}. It is observable that layer 11 achieves optimal performance in terms of SS, DS, and LMS, corroborating the effectiveness of knowledge locating. 
Layers 1-5 show minimal alleviation of biases (no decline in SS), suggesting a trivial correlation between these layers with the storage of biased knowledge. Notably, layers 6-10 not only result in a reduction in SS but also a significant decrease in DS, indicating the entanglement of biased knowledge with other knowledge.

\subsection{Ablation Study on Number of External Parameters}

In this section, we verify the robustness of FAST under limited memory sizes. 
We alter the dimension of hidden states (dim) in our FAST, thereby changing the number of external parameters. 
The results are shown in Figure~\ref{Ablation on the Number of External Parameters}. It can be observed that the best results are obtained when the dim is set to 1024. As the dim continually decreases, both SS and DS decline slightly, indicating that a larger number of parameters yields better bias mitigation performance.
Further increases in dim do not yield better debiasing results. 
Therefore, we decide 1024 to be the dim.

\subsection{Computational Complexity Analysis}
In Table~\ref{Computational Complexity Analysis}, we report the number of parameters and operation time of our proposed FAST on the largest and smallest models in our experiments. The time is counted on a single RTX 3090 with one biased knowledge. 
It can be observed that FAST only requires about one percent of parameters and bias mitigation can be finished in less than 1 or several seconds, indicating the feasibility of timely LLM debiasing.

\begin{table}[htb]
\centering
\scalebox{0.85}{
\begin{tabular}{l  | c  c  c }
\toprule
\textbf{Stage} &  $\textbf{Params}_{\textbf{ Total}}$ & $\textbf{Params}_{\textbf{ FAST}}$ & $\textbf{Time}$  \\
\midrule
{\textit{BERT}}\\
 \textbf{Step~1} &  - & - & 0.83s \\

 \textbf{Step~2} &  0.11B & 0.0016B & 0.66s  \\
 \midrule
{\textit{Llama-2}}\\
 \textbf{Step~1} &  - & - & 24.57s  \\
\textbf{Step~2}&  6.82B & 0.09B & 7.82s \\
\bottomrule
\end{tabular}
}
\caption{Computational complexity analysis on BERT and Llama-2. ``B'' is the abbreviation for billion.}
\label{Computational Complexity Analysis}
\end{table}

\section{Related Works}
\subsection{PLM Debiasing}
Several approaches have been proposed for debiasing pre-trained language models.
The techniques can be grouped into two categories. \textit{(1) Fine-tuning}: This branch includes additional pre-training on re-balanced corpus~\cite{zmigrod2019counterfactual, webster2020measuring} or with a contrastive objective~\cite{he2022mabel, cheng2021fairfil}, projection-based methods~\cite{liang2020towards, ravfogel2020null, kaneko2021debiasing, dev2020measuring} in the embedding space, in-training-based methods~\cite{han2021diverse, he2022mabel} and parameter-efficient fine-tuning~\cite{lauscher2021sustainable, xie2023empirical} methods. 
\textit{(2) Prompt-tuning}: Prompt-tuning~\cite{guo2022auto, yang2023adept, li2023prompt, dong2023co} involves the generation of either discrete prompts or continuous prompts to mitigate social biases. There are also \textit{post-hoc} approaches~\cite{schick2021self} which are deployed in the inference phase.
However, existing techniques treat social groups as interchangeable~\cite{gallegos2023bias}. They seek to neutralize all protected attributes in the inputs or outputs of a model. These strategies tend to ignore or conceal distinct mechanisms of different social groups~\cite{hanna2020towards}. In this paper, we develop evaluation and mitigation strategies that target specific historical biases, without defaulting to the erasure of social group identities as an adequate debiasing strategy.

\subsection{Knowledge Locating}
\label{Knowledge Locating}

Localization aims to interpret a specific model component, including neurons, layers, or subnetworks~\cite{elhage2021mathematical, rogers2021primer, schneider2021explaining, zeiler2014visualizing, wang2022finding, bolukbasi2021interpretability, chen2024learnable}. For example, \cite{dai2021knowledge} identifies a small set of knowledge neurons for each relational fact in BERT. \cite{meng2022locating} locate relational facts to middle FFN layers in autoregressive LLMs, specifically when the model processes the last token of the subject. In contemporaneous work, \cite{bayazit2023discovering} proposes a differentiable weight masking method to discover sparse subnetwork in GPT2 responsible for specific knowledge. 
\cite{meng2022locating} and \cite{meng2022mass} propose to utilize causal mediation analysis~\cite{vig2020investigating, finlayson2021causal} to identify individual layers and neurons that contribute to knowledge storing. Extending these ideas, we propose a two-step model-debiasing framework which firstly locates the key component responsible for storing specific biased knowledge and then calibrates the biased predictions.

\subsection{Model Editing}

Model Editing~\cite{sinitsin2020editable, de2021editing} has been proposed to facilitate data-efficient modifications to model behavior while ensuring no detrimental impact on performance across other inputs. These approaches manipulate the model's output for specific cases either by integrating external models with the original, unchanged model~\cite{mitchell2022memory, murty2022fixing, dong2022calibrating, hartvigsen2022aging, huang2023transformer, zheng2023can} or by altering the model parameters responsible for undesirable output~\cite{mitchell2021fast, dai2021knowledge, li2023pmet, gupta2023editing, hase2021language, meng2022locating}. The most relevant line of works in this regard is locate and edit~\cite{meng2022locating,meng2022mass,dai2021knowledge, li2023pmet}, which suggests identifying neuron activations crucial to a model's factual predictions and subsequently updating the feed-forward weights to edit the output. To the best of our knowledge, we are the first to use model editing techniques to achieve fine-grained model debiasing.

\section{Limitation and Future Works}
\label{Limitation and Future Works}

While our research yields important contributions, we acknowledge the presence of certain limitations.
Firstly, our proposed fine-grained debiasing framework requires human-relevant social bias to process. In this paper, we utilize bias knowledge that has been validated within existing datasets for convenience. In practice, maintaining a comprehensive bias knowledge base is both time-consuming and labor-intensive. We notice that recent works~\citep{sahoo2022detecting, dev2023building} have proposed an automated social bias detection method. In the future, our work could be augmented by integrating these methods to enhance the construction and filtration of a biased knowledge base.
Besides, expanding our fairness edit method against attack scenarios constitutes one of our future research endeavors~\cite{lyu2022study, lyu2023attention}. Finally, compared to the results on BERT and GPT2, the debiasing performance on larger models (Section~\ref{Experiment}) appears less pronounced. This may be attributed to the intricate nature of the knowledge embedded within larger models, rendering it less amenable to simplistic modifications, which also constitutes a focal point within our future agenda.

\end{document}

%% file: math_commands.tex
\usepackage{amsmath,amsfonts,bm}

\def\eqref#1{equation~\ref{#1}}

\def\1{\bm{1}}

\DeclareMathAlphabet{\mathsfit}{\encodingdefault}{\sfdefault}{m}{sl}
\SetMathAlphabet{\mathsfit}{bold}{\encodingdefault}{\sfdefault}{bx}{n}

